\documentclass[10pt,twocolumn,letterpaper]{article}

\usepackage{cvpr}
\usepackage{times}
\usepackage{epsfig}
\usepackage{graphicx}
\usepackage{amsmath}
\usepackage{amssymb}
\usepackage{multirow}
\usepackage{bbm}
\usepackage{subfigure}
\usepackage{booktabs}

\usepackage{comment}
\usepackage{algpseudocode}
\usepackage{makecell}

\usepackage[breaklinks=true,bookmarks=false]{hyperref}

\cvprfinalcopy 

\def\cvprPaperID{9458} 

\begin{document}

\title{Single-Side Domain Generalization for Face Anti-Spoofing}

\author{Yunpei Jia\textsuperscript{\rm 1,2}, Jie Zhang\textsuperscript{\rm 1,2}, Shiguang Shan\textsuperscript{\rm 1,2,3}, Xilin Chen\textsuperscript{\rm 1,2}\\
       \textsuperscript{\rm 1}Key Lab of Intelligent Information Processing of Chinese Academy of Sciences (CAS),\\Institute of Computing Technology, CAS, Beijing, 100190, China\\
       \textsuperscript{\rm 2}University of Chinese Academy of Sciences, Beijing, 100049, China\\
       \textsuperscript{\rm 3}CAS Center for Excellence in Brain Science and Intelligence Technology, Shanghai, 200031, China\\
       {\tt\small yunpei.jia@vipl.ict.ac.cn, \{zhangjie, sgshan, xlchen\}@ict.ac.cn}
}

\maketitle
\thispagestyle{empty}

\begin{abstract}
Existing domain generalization methods for face anti-spoofing endeavor to extract common differentiation features to improve the generalization. However, due to large distribution discrepancies among fake faces of different domains, it is difficult to seek a compact and generalized feature space for the fake faces. In this work, we propose an end-to-end single-side domain generalization framework (SSDG) to improve the generalization ability of face anti-spoofing. The main idea is to learn a generalized feature space, where the feature distribution of the real faces is compact while that of the fake ones is dispersed among domains but compact within each domain. Specifically, a feature generator is trained to make only the real faces from different domains undistinguishable, but not for the fake ones, thus forming a single-side adversarial learning. Moreover, an asymmetric triplet loss is designed to constrain the fake faces of different domains separated while the real ones aggregated. The above two points are integrated into a unified framework in an end-to-end training manner, resulting in a more generalized class boundary, especially good for samples from novel domains. Feature and weight normalization is incorporated to further improve the generalization ability. Extensive experiments show that our proposed approach is effective and outperforms the state-of-the-art methods on four public databases. The code is released online\footnote{https://github.com/taylover-pei/SSDG-CVPR2020}.
\end{abstract}

\section{Introduction}
\begin{figure}[!t]
    \centering
    \includegraphics[scale=0.7]{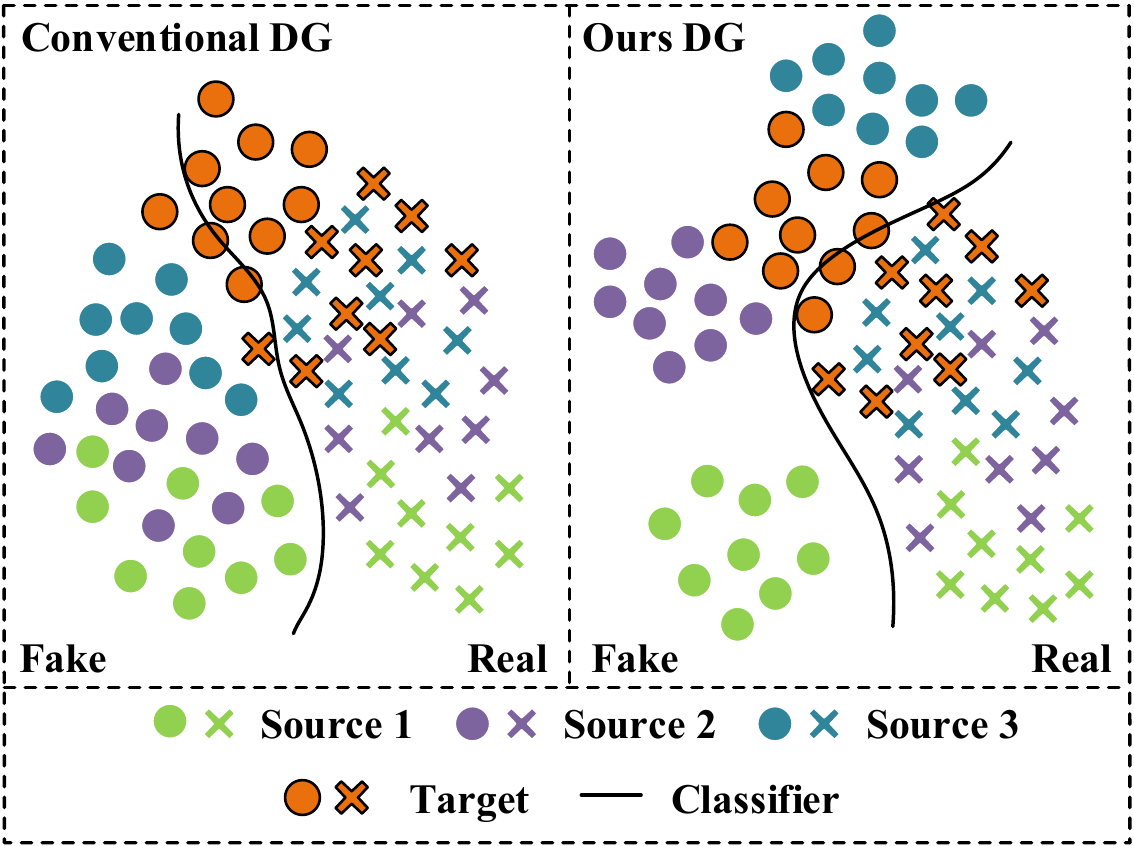}
    \caption{Left: Conventional domain generalization methods align source domains to learn a shared feature space, which fail to get a discriminative class boundary on the unseen domain. Right: Our single-side domain generalization method aggregates all the real examples while separates the fake ones from different domains to learn a class boundary, generalizing better to the novel domain.}
    \label{fig:motivation}
\end{figure}
In recent years, face recognition techniques have been widely exploited in our daily life, especially in the fields of smartphones login, access control, etc.
However, many presentation attacks have emerged (\emph{e.g.}, print attack, video attack, and 3D mask attack), which has led to a huge security risk on face recognition systems and become an increasingly critical concern in the face recognition field.
To tackle this issue, various face anti-spoofing methods have been proposed, which can be coarsely categorized into texture-based methods and temporal-based methods.
The texture-based methods utilize hand-craft descriptors or data-driven deep learning to extract texture cues discriminative between the real faces and the fake ones, such as the color \cite{boulkenafet2016face}, distortion cues \cite{galbally2013image, wen2015face}, etc.
In contrast, the temporal-based methods leverage various temporal cues in consecutive face frames, such as rPPG \cite{liu20163d, liu2018remote} and optical flow \cite{anjos2013motion,bao2009liveness}. 

Although existing state-of-the-art methods have obtained promising results under intra-database testing scenarios, they cannot generalize well in case of cross-database testing, where training (source domain\footnote{The term domain in this paper represents a database.}) and testing (target domain) data come from different domains.
The reason behind is that traditional methods take no consideration of the intrinsic distribution relationship among different domains and thus extract discriminative features of database biased \cite{torralba2011unbiased}, leading to poor generalization to unseen domains.
To address the problem, recent face anti-spoofing methods \cite{li2018unsupervised, wangimproving} adopt domain adaptation techniques to minimize the distribution discrepancy between the source and the target domain by utilizing unlabeled target data.
However, in many real-world scenarios, it is difficult and expensive to collect a lot of unlabeled target data for training, and even no information about the target domain is available.

Therefore, some researchers start to address the face anti-spoofing problem from the perspective of domain generalization (DG), which aims to train a model by utilizing multiple existing source domains without seeing any target data.
Conventional DG approaches \cite{li2018domain, shao2019multi} aim to learn a generalized feature space by aligning the distributions among multiple source domains.
And they assume that the extracted features of unseen faces can be mapped nearby the shared feature space so that the model can generalize well to the novel domains.
Since the real faces from both the source and the target domains are collected by imaging real people, their distribution discrepancies are small, which makes it relatively easy to learn a compact feature space.
In contrast, due to the diversity of attack types and collecting ways, it is relatively hard to aggregate the features of fake faces from different domains together.
Therefore, seeking a generalized feature space for the fake faces is difficult to optimize and may also affect the classification accuracy for the target domain \cite{akuzawa2019adversarial, xie2017controllable}.
For this reason, as illustrated in the left of Figure \ref{fig:motivation}, although a compact feature space for both the real and the fake examples is achieved, it still fails to learn a discriminative class boundary for the novel target domain.
In consideration of the above arguments, besides constraining the real faces and the fake ones to be as distinguishable as possible, we propose to pull all the real faces aggregated while push the fake ones of different domains separated.
As illustrated in the right of Figure \ref{fig:motivation}, our method aims at forcing the features of fake faces more dispersed in the feature space while those of the real ones more compact, thus leading to a class boundary, which generalizes better to the target domain.

\begin{figure*}[!t]
    \centering
    \includegraphics[scale=0.93]{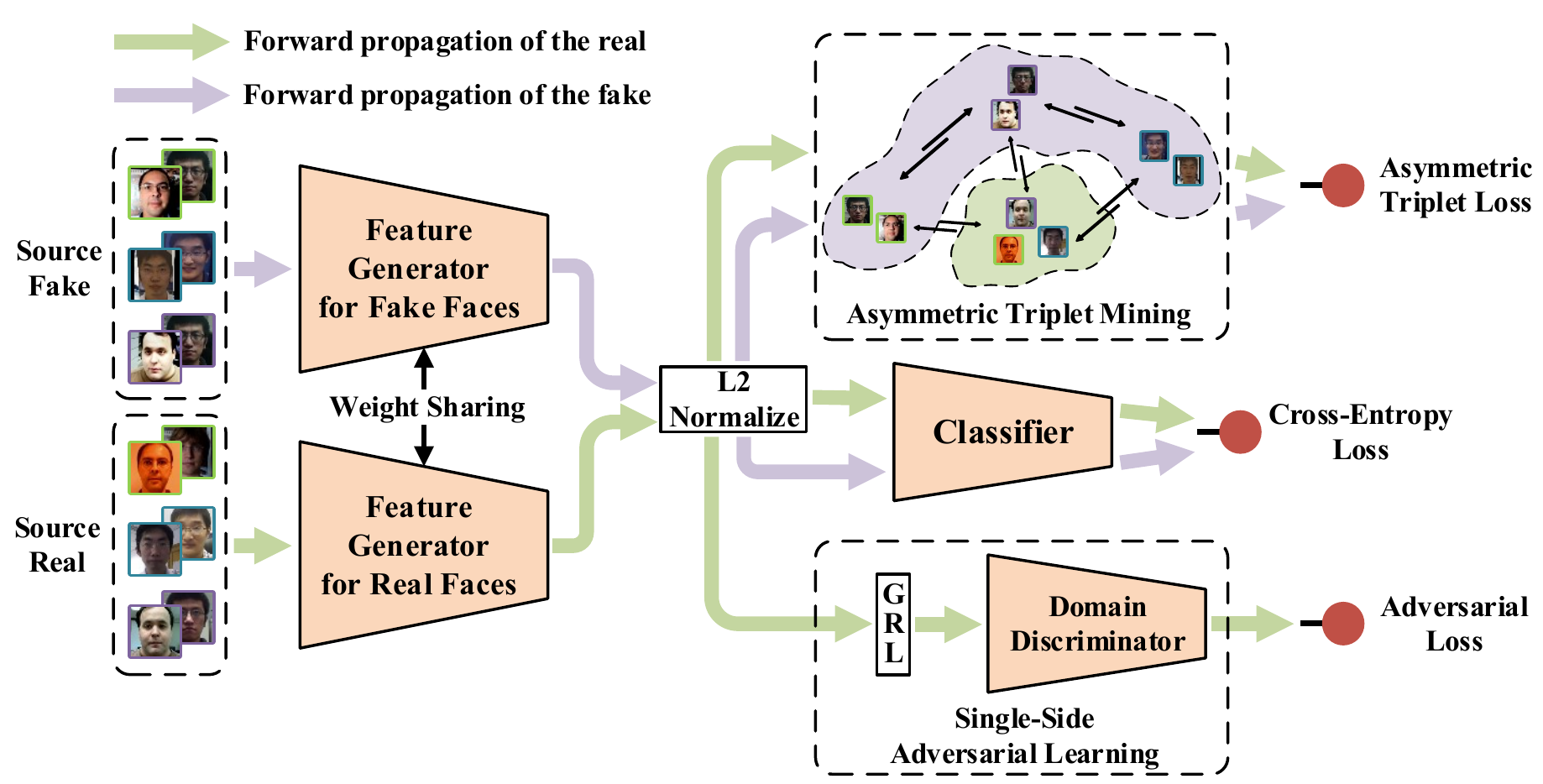}
    \caption{An overview of the proposed method. The input faces with different color borders represent examples of different domains. The parameter sharing feature generator is trained to make the feature distributions of different domains undistinguishable for the real faces but not for the fake ones under the single-side adversarial learning. Moreover, the asymmetric triplet mining is implemented to separate the fake faces while aggregate the real ones of different domains to force the features of fake faces to be more dispersed in the feature space. The feature and weight normalization is incorporated to further improve the generalization ability.}
    \label{fig:architecture}
\end{figure*}

With the above thoughts in mind, we propose an end-to-end single-side domain generalization framework (SSDG), as shown in Figure \ref{fig:architecture}.
Specifically, a feature generator is trained competing with a domain discriminator to make the features of real faces from different domains undistinguishable, forming a single-side adversarial learning. 
Since the fake faces are rather diverse than the real ones, we treat the fake faces of different domains as different categories while the real ones of all domains as the other category to perform the asymmetric triplet mining, which ensures three properties: 1) fake faces of different categories are separated; 2) all the real ones regardless of domains are aggregated; 3) all the real faces and the fake ones are distinguishable.
As a result, two feature distributions with different characteristics can be achieved, leading to a better generalized class boundary for the target domains.
Meanwhile, feature and weight normalization is incorporated to further improve the generalization ability during training.

The main contributions of this work are summarized as follows: 1) Based on the analysis that the fake faces are rather diverse than the real ones, we propose a novel end-to-end single-side domain generalization framework. 2) We design the single-side adversarial learning and the asymmetric triplet loss to achieve different optimization goals for the real and the fake faces and perform the feature and weight normalization to further improve the performance. 3) We make comprehensive comparisons and achieve the state-of-the-art performance on four public databases.

\section{Related Work}
\subsection{Face Anti-spoofing Methods}
In this subsection, we review the most representative face anti-spoofing methods, which can be generally divided into two groups: texture-based methods and temporal-based methods, as already mentioned previously. 

\textbf{Texture-based methods} distinguish the real faces from the fake ones through various texture cues.
Many prior works adopt hand-craft descriptors for face anti-spoofing, such as LBP \cite{de2014face, maatta2011face}, HOG \cite{gragnaniello2015investigation}, SURF \cite{boulkenafet2016face}, SIFT \cite{patel2016secure}, etc. 
In recent years, with the rapid development of deep learning in computer vision, various methods turn to employ CNNs to extract more discriminative features.
Yang \emph{et al.} \cite{yang2014learn} are the first to use CNN with binary supervision for face anti-spoofing.
Atoum \emph{et al.} \cite{atoum2017face} propose a two-stream CNN architecture to extract depth features combining with the texture features to detect attacks.
And the face de-spoofing method \cite{jourabloo2018face} inversely decomposes a spoof face into a live face and a spoof noise for classification.

\textbf{Temporal-based methods} make use of temporal cues in consecutive face frames for spoofing face detection.
Mouth-motion detection \cite{kollreider2007real} and eye-blinking detection \cite{pan2007eyeblink, sun2007blinking} are among the earliest solutions for face anti-spoofing based on the temporal cues. 
Recently, there exist more general methods relying on more effective temporal cues, instead of the particular liveness information. 
CNN-LSTM architecture is proposed in \cite{xu2015learning} to take multiple frames as input to extract temporal features for face anti-spoofing.
Liu \emph{et al.} \cite{liu2018learning} utilize the rPPG signal as the auxiliary supervision with a novel CNN-RNN network to detect attacks.
More robust rPPG features are extracted by \cite{liu20163d, liu2018remote} to detect 3D mask attack effectively.
And Yang \emph{et al.} \cite{yang2019face} take into consideration the global temporal and local spatial cues to distinguish the real faces from the fake ones.

Although the above methods have obtained remarkable results under intra-database testing scenarios, they cannot mine the distribution relationship among different domains and might suffer from extracting database-biased features, leading to poor generalization to unseen domains.

\subsection{Domain Generalization}
Both the domain adaptation methods \cite{li2018unsupervised, tu2019deep, wangimproving} and zero-shot face anti-spoofing methods \cite{liu2019deep, qin2019learning} aim to improve the generalization ability. 
In contrast, the domain generalization (DG) methods explicitly mine the relationship among multiple source domains without accessing any target data, which generalize better to unseen domains.
Most of the previous DG methods focus on minimizing the distribution discrepancies among multiple source domains to extract domain-invariant features.    
Motiian \emph{et al.} \cite{motiian2017unified} propose a new loss to guide the features of the same class to be close in the latent feature space.
Autoencoders are exploited in \cite{ghifary2015domain, li2018domain} to align the distributions of source domains for generalized features.
The most related work to ours is proposed in \cite{shao2019multi}, where multiple feature extractors are trained to learn a generalized feature space via adversarial learning.
However, as the first attempt to address the face anti-spoofing problem from the DG point of view, its training process is not end-to-end.
Moreover, due to the diversity of attack types and collecting ways, it is difficult to seek a generalized feature space for the fake faces, usually leading to a sub-optimal solution for face anti-spoofing.

\begin{figure*}[!t]
    \centering
    \includegraphics[scale=0.35]{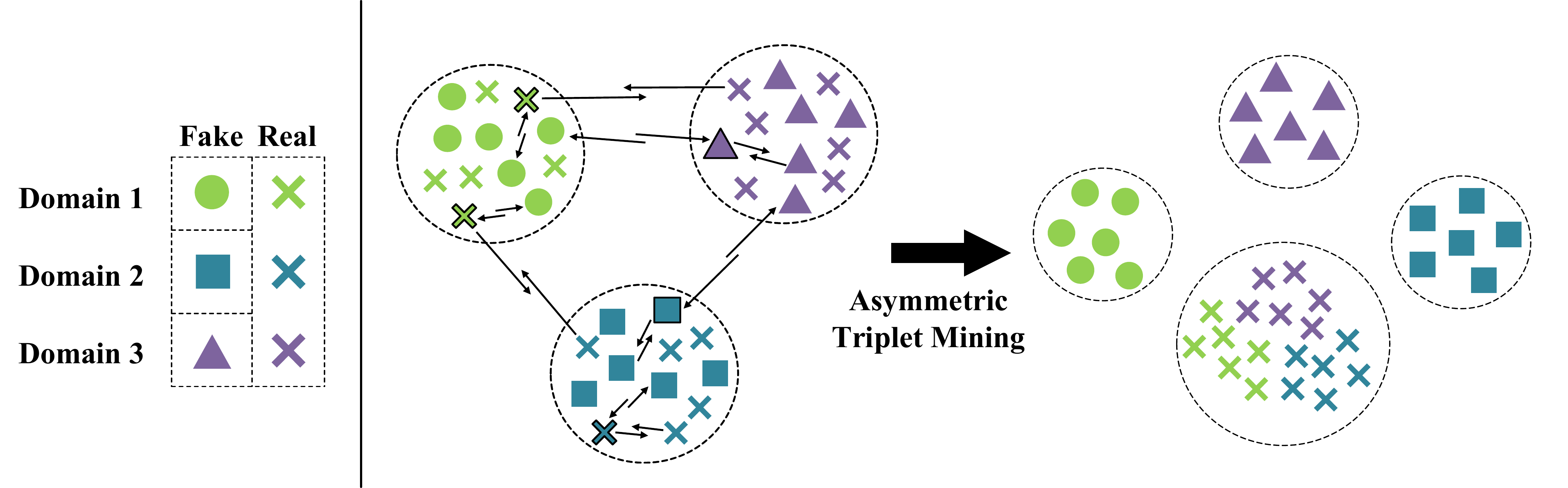}
    \caption{Illustration of the asymmetric triplet loss. A shape with the black border represents the anchor point, while the other two points linked with it are the positive and the negative ones, respectively. Asymmetric triplet mining is conducted to separate the fake faces of different domains while aggregate the real ones together. Meanwhile, the fake faces are also pulled apart away from all the real ones. After adopting the asymmetric triplet loss, the features of fake faces can be more dispersed in the feature space, leading to a class boundary better generalized.}
    \label{fig:triplet}
\end{figure*}
\section{Proposed Method}
\subsection{Overview}
\label{section:Overview}
Since the distribution discrepancies are much larger among the fake faces than the real ones, it is nontrivial to align the features of fake faces from different domains. 
Therefore, seeking a compact and generalized feature space for both the real and the fake faces is difficult to optimize and may bring negative influences on the classification accuracy for unseen domains.
In this work, we focus on asymmetric optimization goals for the real and the fake faces belonging to different domains to learn a feature space with higher generalization ability to unseen domains.
As illustrated in Figure \ref{fig:architecture}, we propose a single-side domain generalization framework (SSDG) for face anti-spoofing.
Specifically, the feature generator is trained competing with the domain discriminator to make the features of real faces undistinguishable, forming a single-side adversarial learning process.
Moreover, we propose the asymmetric triplet loss to explicitly separate the fake faces of different domains while aggregate the real ones.
Additionally, feature and weight normalization is further incorporated to improve the generalization ability during the training process.
Therefore, the proposed SSDG method forces the fake faces to be more dispersed in the feature space while the real ones to be more compact, leading to a more generalized class boundary to unseen domains.

\subsection{Single-Side Adversarial Learning}
Assume there are $N$ source domains, denoted as $D = \{D_1, D_2, ..., D_N\}$.
Each of them contains two categories of face images, \emph{i.e.}, the real faces $X_r$ and the fake faces $X_f$.
Since all the real faces are collected by imaging real people, we conjecture that the distribution discrepancies among them are much smaller compared to the fake ones.
Therefore, seeking a generalized feature space for the real faces is relatively easy, which promotes to capture more common discriminative cues.
Specifically, we propose the single-side adversarial learning to learn a generalized feature space, which is conducted only on the extracted features of real faces.
In contrast, the adversarial learning is not performed for the fake ones.

We firstly separate the real faces from the fake ones of all source domains, and then feed them into the corresponding feature generators, which transform the input faces into a latent feature space as follows:
\begin{equation}
    Z_r = G_r(X_r), Z_f = G_f(X_f),
\end{equation}
where $G_r$, $G_f$ represent the feature generators for the real and the fake faces, respectively, and $Z_r$, $Z_f$ are the corresponding extracted features. 
Since a parameter sharing strategy is adopted to make all the parameters of $G_r$ and $G_f$ identical, we refer them collectively as $G$ in the following for the sake of convenience. 
The domain discriminator, denoted as $D$, is implemented based on $Z_r$ to determine which source domain the input features stem from.
On the contrary, the feature generator is trained to spoof the domain discriminator so that the domain labels cannot be recognized.
Therefore, a single-side adversarial learning procedure is designed between the feature generator and the domain discriminator to learn a generalized feature space for the real faces.
During the learning procedure, the parameters of feature generator are optimized by maximizing the loss of domain discriminator while those of domain discriminator are optimized with the opposite objective.
Since there are multiple source domains for classification, we utilize the standard cross-entropy loss to optimize the network under the single-side adversarial learning:
\begin{equation}
\begin{aligned}
    \min_{D}\max_{G} &\mathcal{L}_{Ada}(G,D) = \\
    &-\mathbb{E}_{x, y \sim X_r,Y_D} \sum\nolimits_{n=1}^{N} \mathbbm{1}_{[n=y]} \log D(G(x)),
\end{aligned}
\end{equation}
where $Y_D$ represents the set of domain labels.

In order to optimize the feature generator and the domain discriminator simultaneously, a gradient reverse layer (GRL) \cite{ganin2014unsupervised} is inserted after the feature generator, which multiplies the gradient of the adversarial loss by $-\lambda$ during backward propagation.
We set $\lambda = \frac{2}{1+\exp{(-10k)}}-1$ and $k = \frac{current\_iters}{total\_iters}$ with the same purpose introduced by \cite{ganin2014unsupervised} to suppress the effect of the noisy signals at the early training stage.
With the single-side adversarial learning, a generalized feature space for the real faces is achieved, where common discriminative cues can be further exploited.

\subsection{Asymmetric Triplet Mining}
Due to the diversity of attack types and database collection ways, the distribution discrepancies are much larger among the fake faces than the real ones.
Therefore, seeking a dispersed feature space for the fake is relatively easy compared to seeking a compact one.
In consideration of this, we explicitly separate the fake faces of different domains to force them to be more dispersed in the feature space.
In contrast, we aggregate all the real ones to force them to be more compact.
To achieve the asymmetric optimization goals for the real and fake faces, we propose the asymmetric triplet loss to perform the asymmetric triplet mining according to the categories, which promotes to learn a better class boundary for unseen domains.

Specifically, assuming there are three source domains available, we recombine the real and the fake faces coming from three different domains into four categories.
As shown in the left of Figure \ref{fig:triplet}, the fake faces of three different domains are treated as distinct categories (circle, square, and triangle, respectively), while all the real ones are put together into one category (cross).
And then, four-category asymmetric triplet mining is conducted on the real and the fake faces to achieve the following optimization goals: 1) separate the fake faces of different domains; 2) aggregate the real faces of all source domains; 3) pull apart the fake faces away from all the real ones.
After that, as shown in the right of Figure \ref{fig:triplet}, the extracted features of fake faces are more dispersed than before in the feature space and those of real ones are more aggregated, leading to a better generalized class boundary for unseen domains.
The feature generator is optimized as follows:
\begin{equation}
\begin{aligned}
    \min_{G} \mathcal{L}_{AsTrip}(G) &= \sum_{x_{i}^{a}, x_{i}^{p}, x_{i}^{n}} (\left\|f\left(x_{i}^{a}\right)-f\left(x_{i}^{p}\right)\right\|_{2}^{2} \\ &- \left\|f\left(x_{i}^{a}\right)-f\left(x_{i}^{n}\right)\right\|_{2}^{2}+\alpha ),
\end{aligned}
\end{equation}
where the labels of anchor $x_i^a$ and positive example $x_i^p$ are the same, while those of $x_i^a$ and negative example $x_i^n$ are different. The $\alpha$ is a pre-defined margin.

\subsection{Feature and Weight Normalization}
Normalization approaches have been verified effective in the field of face recognition.
In this work, both feature normalization and weight normalization are incorporated to further improve the generalization ability of the proposed method.

\textbf{Feature Normalization.} 
The feature norms are highly related to the quality of the images, as discussed in \cite{ranjan2017l2, wang2018additive}.
Due to the diversity of database collecting conditions (\emph{e.g.}, illustration, camera quality, etc.), large differences exist among the feature norms of different face images under both the intra-database and cross-database scenarios, which hinder the feature learning process and also affect the generalization ability.
Thus, we perform the $l_2$ normalization on the outputs of the feature generator to constrain all the features share the same Euclidean norm to further improve the performance of face anti-spoofing.

\textbf{Weight Normalization.} 
In this work, the face anti-spoofing problem is regarded as a binary classification task. 
Since the softmax function is utilized for training, the decision boundary can be achieved between the real and the fake faces as $ \|\boldsymbol{W}_{1}^{T}\|\|\tilde{\mathbf{z}}\| \cos(\theta_{1})+b_{1}=\|\boldsymbol{W}_{0}^{T}\|\|\tilde{\mathbf{z}}\| \cos(\theta_{0})+b_{0}$, 
where $\boldsymbol{W_i}$ is the $i$-th column of the parameter matrix in last fully connected layer, $b_i$ is the corresponding bias, and $\theta_i$ is the angle between the normalized feature $\tilde{\mathbf{z}}$ and $\boldsymbol{W_i}$.
Following the works of \cite{deng2019arcface, liu2017sphereface, wang2018additive}, we perform $l_2$ normalization on $\boldsymbol{W_i}$ to fix $\|\boldsymbol{W}_i\|=1$ and set $b_i=0$, which makes the decision boundary becomes $\cos(\theta_{1})-\cos(\theta_{0})=0$.
Therefore, we further constrain the feature learning process by the weight normalization, which promotes to learn more discriminative cues between the real and the fake faces.

\begin{table*}[]
    \centering
    \caption{Evaluations of different components of the proposed method with different architectures.}
    \begin{tabular}{c|c|c|c|c|c|c|c|c}
        \toprule
        \multirow{2}*{\textbf{Method}} & \multicolumn{2}{c|}{\textbf{O\&C\&I to M}} & \multicolumn{2}{c|}{\textbf{O\&M\&I to C}} & \multicolumn{2}{c|}{\textbf{O\&C\&M to I}} & \multicolumn{2}{c}{\textbf{I\&C\&M to O}} \\
        \cline{2-9}
        ~ & HTER(\%) & AUC(\%) & HTER(\%) & AUC(\%) & HTER(\%) & AUC(\%) & HTER(\%) & AUC(\%)  \\
        \hline
        \hline
        SSDG-M w/o triplet   & 21.19 & 83.54 & 26.78 & 79.10 & 23.93 & 74.86 & 25.43 & 80.52 \\
        SSDG-M w/o ssad  & 24.05 & 81.94 & 28.11 & 80.15 & 21.29 & 84.52 & 26.62 & 79.59 \\
        SSDG-M w/o norm      & 17.86 & 89.76 & 30.11 & 78.38 & 25.57 & 73.92 & 29.74 & 75.48 \\
        \textbf{SSDG-M} & \textbf{16.67}  & \textbf{90.47} & \textbf{23.11} & \textbf{85.45} & \textbf{18.21} & \textbf{94.61} & \textbf{25.17} & \textbf{81.83} \\
        \hline
        \hline
        SSDG-R w/o triplet   &  8.81 & 96.85 & 14.33 & 92.28 & 15.21 & 83.09 & 21.98 & 85.54 \\
        SSDG-R w/o ssad  & 11.19 & 95.10 & 12.89 & 94.08 & 12.14 & \textbf{96.63} & 18.06 & 90.43 \\
        SSDG-R w/o norm      & 10.24 & 96.58 & 12.78 & 95.06 & 12.64 & 92.92 & 15.99 & 91.26 \\
        \textbf{SSDG-R} & \textbf{7.38}  & \textbf{97.17} & \textbf{10.44} & \textbf{95.94} & \textbf{11.71} & 96.59 & \textbf{15.61} & \textbf{91.54} \\
        \bottomrule
    \end{tabular}
    \label{tab:component_ablasion}
\end{table*}
\begin{table*}[]
    \centering
    \caption{Comparison results between the proposed method and the corresponding baseline method with different architectures.}
    \begin{tabular}{c|c|c|c|c|c|c|c|c}
        \toprule
        \multirow{2}*{\textbf{Method}} & \multicolumn{2}{c|}{\textbf{O\&C\&I to M}} & \multicolumn{2}{c|}{\textbf{O\&M\&I to C}} & \multicolumn{2}{c|}{\textbf{O\&C\&M to I}} & \multicolumn{2}{c}{\textbf{I\&C\&M to O}} \\
        \cline{2-9}
        ~ & HTER(\%) & AUC(\%) & HTER(\%) & AUC(\%) & HTER(\%) & AUC(\%) & HTER(\%) & AUC(\%)  \\
        \hline
        \hline
         BDG-M & 17.14 & 87.70 & 28.00 & 73.42 & 20.93 & 87.06 & 26.27 & 79.99 \\
        \textbf{SSDG-M} & \textbf{16.67}  & \textbf{90.47} & \textbf{23.11} & \textbf{85.45} & \textbf{18.21} & \textbf{94.61} & \textbf{25.17} & \textbf{81.83} \\
        \hline
        \hline
        BDG-R &  9.52 & 93.52 & 12.78 & 94.38 & 12.86 & 93.06 & 16.46 & 91.39 \\
        \textbf{SSDG-R} & \textbf{7.38}  & \textbf{97.17} & \textbf{10.44} & \textbf{95.94} & \textbf{11.71} & \textbf{96.59} & \textbf{15.61} & \textbf{91.54} \\
        \bottomrule
    \end{tabular}
    \label{tab:baseline_ablasion}
\end{table*}

\subsection{Loss Function}
Since all the source domain data contain labels, a face anti-spoofing classifier is implemented after the feature generator, as illustrated in Figure \ref{fig:architecture}.
Both the face anti-spoofing classifier and the feature generator are optimized by the standard cross-entropy loss, denoted as $\mathcal{L}_{Cls}$.
Integrating all things mentioned above together, the objective of the proposed single-side domain generalization framework for face anti-spoofing is:
\begin{equation}
    \mathcal{L}_{SSDG} = \mathcal{L}_{Cls} + \lambda_1 \mathcal{L}_{Ada} + \lambda_2 \mathcal{L}_{AsTrip},
\end{equation}
where $\lambda_1$ and $\lambda_2$ are the balanced parameters.
Instead of decomposing the training process into two phases in \cite{shao2019multi}, we train all the components in an end-to-end manner. 

\section{Experiment}
\subsection{Experimental Settings}
\textbf{Databases.} Four public face anti-spoofing databases are utilized to evaluate the effectiveness of our method: OULU-NPU \cite{boulkenafet2017oulu} (denoted as O), CASIA-FASD \cite{Zhang2012AFA} (denoted as C), Idiap Replay-Attack \cite{chingovska2012effectiveness} (denoted as I), and MSU-MFSD \cite{wen2015face} (denoted as M).
We randomly select one database as the target domain for testing and the remaining three as the source domains for training. 
Thus, we have four testing tasks in total: O\&C\&I to M, O\&M\&I to C, O\&C\&M to I, and I\&C\&M to O.
Many differences (\emph{e.g.}, background, resolution, illustration, ethnicity, etc.) exist under both intra-database and cross-database testing scenarios, especially for the fake ones, which cause great distribution discrepancies among them.

\textbf{Implementation Details.} 
MTCNN algorithm \cite{zhang2016joint} is adopted for face detection and face alignment to perform the data pre-processing.
All the detected faces are normalized to 256$\times$256$\times$3 as the input of the network, where only RGB channels are utilized for training to further reduce the network complexity.
We train our model using only one frame information randomly selected from each video.
The SGD optimizer with momentum of 0.9 and weight decay of 5e-4 is used for the optimization.
The hyperparameter $\alpha$ is set to 0.1.

Our framework is implemented by PyTorch. And two different architectures of the feature generator are adopted for comparisons.
The first one combines the feature generator with the feature embedder defined in MADDG \cite{shao2019multi}.
For the second one, we replace the last average pooling layer of ResNet-18 \cite{he2016deep} by the global pooling layer (GAP) and utilize all the above layers of GAP. 
Specifically, we add a fully connected layer (FC) as the bottleneck layer on top of each feature generator, which consists of 512 hidden units.
The face anti-spoofing classifier is a simple linear model with a 2 nodes FC layer.
And the domain discriminator contains two FC layers with 512 and 3 nodes, respectively.
We denote these two different architectures by M and R for short in the following (\emph{i.e.}, SSDG-M and SSDG-R).

\textbf{Evaluation Metrics.}
Following the work of \cite{shao2019multi}, we use the Half Total Error Rate (HTER) and the Area Under Curve (AUC) as the evaluation metrics.
Moreover, the Receiver Operating Characteristic (ROC) and some visualizations (t-SNE \cite{maaten2008visualizing} and CAM \cite{selvaraju2017grad}) are also reported to further evaluate the performance.

\subsection{Discussion}
\subsubsection{Influences of Each Network Component}
We perform the ablation study to evaluate the performance gained by each component for different network architectures, \emph{i.e.}, the single-side adversarial learning (denoted as ssad), the asymmetric triplet loss (denoted as triplet), and the feature and weight normalization (denoted as norm).
The comparison results are shown in Table \ref{tab:component_ablasion}.

It can be seen that the performances of the proposed method with different architectures both degrade if any component is removed.
The comparison results verify that each component of SSDG contributes to performance improvement and the incorporation of all these components can achieve the best results.

\begin{table*}[]
    \centering
    \caption{Comparison results of two different network architectures of the feature generator.}
    \begin{tabular}{c|c|c|c|c|c}
        \toprule
        Backbones      & Flops(G) & Params(M) & Speed(FPS) & Avg HTER(\%) & Avg AUC(\%) \\
        \hline
        \hline
        MADDG-based    & 47.59 & \textbf{3.35} & 36.40 & 20.79 & 88.09 \\
        ResNet18-based & \textbf{2.38} & 11.18 & \textbf{149.15} & \textbf{11.29} & \textbf{93.81} \\
        \bottomrule
    \end{tabular}
    \vspace{-0.1in}
    \label{tab:backbone_comparison}
\end{table*}

\begin{table}[]
    \centering
    \caption{Comparison results of domain generalization with limited source domains for face anti-spoofing.}
    \begin{tabular}{c|p{1.0cm}<{\centering}|p{1.0cm}<{\centering}|p{1.0cm}<{\centering}|p{1.0cm}<{\centering}}
        \toprule
        \multirow{2}*{\textbf{Method}} & \multicolumn{2}{c|}{\textbf{M\&I to C}} & \multicolumn{2}{c}{\textbf{M\&I to O}} \\
        \cline{2-5}
        ~ & HTER & AUC & HTER & AUC \\
        \hline
        \hline
        MS-LBP \cite{maatta2011face} &  51.16 & 52.09 & 43.63 & 58.07 \\
        IDA \cite{wen2015face} & 45.16 & 58.80 & 54.52 & 42.17 \\
        CT \cite{boulkenafet2016faceLBP} & 55.17 & 46.89 & 53.31 & 45.16 \\
        LBP-TOP \cite{de2014face}  & 45.27 & 54.88 & 47.26 & 50.21 \\ 
        MADDG \cite{shao2019multi} & 41.02 & 64.33 & 39.35 & 65.10 \\
        \hline
        \textbf{SSDG-M} & \textbf{31.89}  & \textbf{71.29} & \textbf{36.01} & \textbf{66.88} \\
        \bottomrule
    \end{tabular}
    \label{tab:limited_domain}
\end{table}

\subsubsection{Comparisons with the Baseline Method}
We further compare the SSDG method with the corresponding baseline method, which aims to seek a generalized feature space for both the real and the fake faces.
Specifically, we add another domain discriminator after the feature generator to perform adversarial learning on both the real and the fake features.
Moreover, the proposed asymmetric triplet loss is replaced with a two-category triplet loss to aggregate all the fake faces as well as the real ones together.
Two different network architectures are adopted in the baseline method for more comparisons (denoted as BDG-M and BDG-R, respectively).
The comparison results are shown in Table \ref{tab:baseline_ablasion}.

Firstly, it can be seen that the performance of BDG-M method is comparable with that of the state-of-the-art MADDG method \cite{shao2019multi} shown in Table \ref{tab:comparison_sota}.
The average HTER results of them for all total testing tasks are 23.09\% and 23.05\%, respectively.
This is because the above two methods both aim to seek a generalized feature space not only for the real faces but also for the fake ones.
In contrast, our SSDG method outperforms the BDG method as well as the MADDG method on all testing tasks with different network architectures, which demonstrates that seeking a generalized feature space for fake faces is sub-optimal.
As a result, it is more feasible for the face anti-spoofing task to apply asymmetric optimization goals for the real and the fake faces, which can get a class boundary, generalizing better to unseen domains.

\begin{figure}[t]
\includegraphics[width=1.0\linewidth]{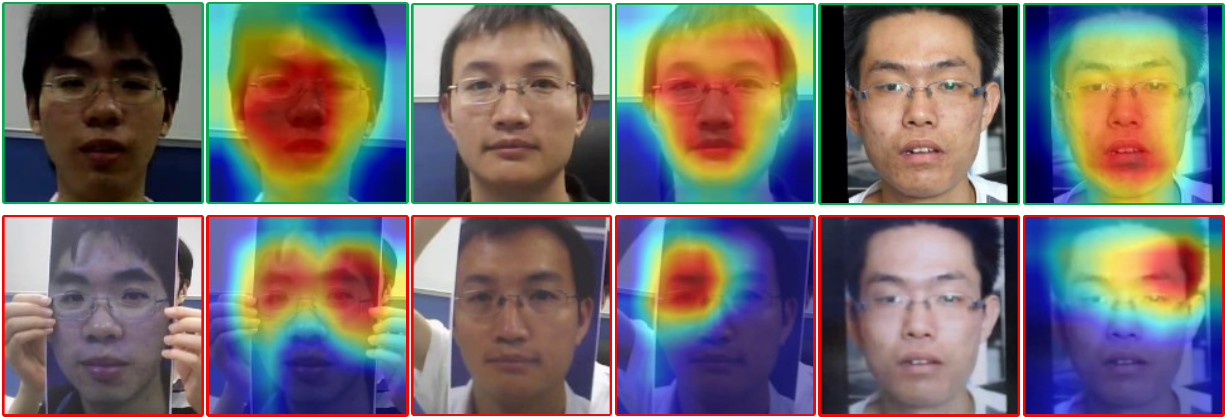}
\caption{Grad-CAM \cite{selvaraju2017grad} visualizations of the SSDG method under O\&M\&I to C. The first row shows the real faces and the second shows the fake ones.}
\label{fig:cam}
\end{figure}
\subsubsection{Visualizations of the Proposed Method}
As shown in Figure \ref{fig:cam}, we adopt the Grad-CAM \cite{selvaraju2017grad} to provide the class activation map (CAM) visualizations of our method. 
It shows that the SSDG method always focuses on the region of the internal face to seek discriminative cues instead of the domain-specific backgrounds, illuminations, etc., which is more likely to generalize well to unseen domains.
Specifically, for the fake faces, our method can pay attention to different regions according to different attacks, such as the eyes region of the face for the cut attack.

\begin{figure}
    \centering
    \hspace{-5mm}
    \subfigure[BDG-R]{
        \includegraphics[scale=.45]{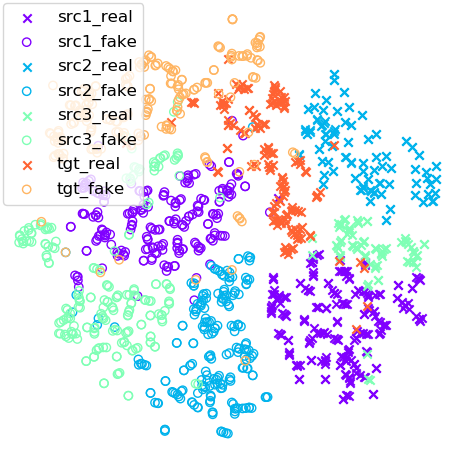}
    }
    \hspace{-3mm}
    \subfigure[SSDG-R]{
        \includegraphics[scale=.45]{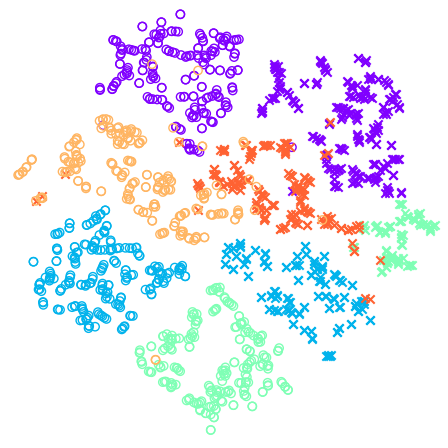}
    }
    \hspace{-5mm}
    \caption{The t-SNE \cite{maaten2008visualizing} visualizations of the extracted features by the BDG-R method (a) and the SSDG-R method (b) under the O\&M\&I to C testing tasks (best viewed in color).}
    \label{fig:tSNE}
\end{figure}

Moreover, as shown in Figure \ref{fig:tSNE}, we randomly select 200 samples of each category from four databases and plot the t-SNE \cite{maaten2008visualizing} visualizations to analyze the feature space learned by the SSDG method and the corresponding baseline BDG method.
It can be seen that the SSDG method can make the features of fake faces more dispersed in the feature space compared to those of the BDG method. 
In contrast, the feature distribution of the real faces is more compact.
Therefore, a better class boundary can be achieved by the SSDG method, which generalizes well to the target domain.

\begin{figure*}[!t]
    \centering
    \vspace{-0.2in}
    \hspace{-2mm}
    \includegraphics[scale=0.31]{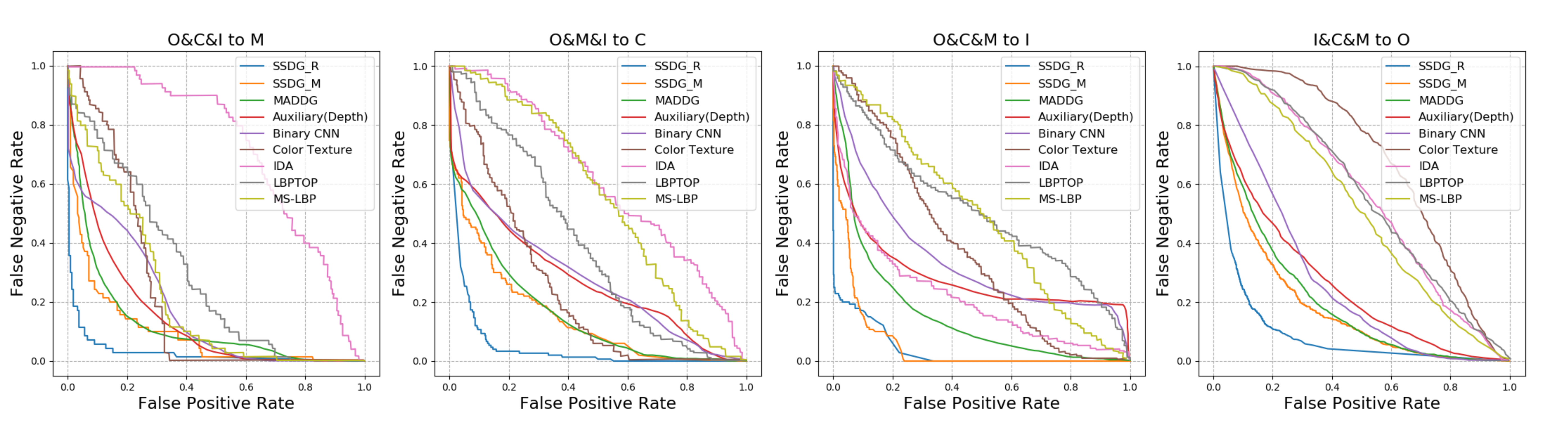}
    \hspace{-2mm}
    \caption{ROC curves of four testing tasks for domain generalization on face anti-spoofing.}
    \label{fig:roc}
\end{figure*}
\begin{table*}[]
    \centering
    \caption{Comparison results between the proposed method and state-of-the-art methods for domain generalization on face anti-spoofing.}
    \begin{tabular}{c|c|c|c|c|c|c|c|c}
        \toprule
        \multirow{2}*{\textbf{Method}} & \multicolumn{2}{c|}{\textbf{O\&C\&I to M}} & \multicolumn{2}{c|}{\textbf{O\&M\&I to C}} & \multicolumn{2}{c|}{\textbf{O\&C\&M to I}} & \multicolumn{2}{c}{\textbf{I\&C\&M to O}} \\
        \cline{2-9}
        ~ & HTER(\%) & AUC(\%) & HTER(\%) & AUC(\%) & HTER(\%) & AUC(\%) & HTER(\%) & AUC(\%)  \\
        \hline
        \hline
        MS-LBP \cite{maatta2011face}         & 29.76 & 78.50 & 54.28 & 44.98 & 50.30 & 51.64 & 50.29 & 49.31 \\
        Binary CNN \cite{yang2014learn}     & 29.25 & 82.87 & 34.88 & 71.94 & 34.47 & 65.88 & 29.61 & 77.54 \\
        IDA \cite{wen2015face}               & 66.67 & 27.86 & 55.17 & 39.05 & 28.35 & 78.25 & 54.20 & 44.59 \\
        Color Texture \cite{boulkenafet2016faceLBP}         & 28.09 & 78.47 & 30.58 & 76.89 & 40.40 & 62.78 & 63.59 & 32.71 \\
        LBP-TOP \cite{de2014face}            & 36.90 & 70.80 & 42.60 & 61.05 & 49.45 & 49.54 & 53.15 & 44.09 \\
        Auxiliary (Depth) & 22.72 & 85.88 & 33.52 & 73.15 & 29.14 & 71.69 & 30.17 & 77.61 \\
        Auxiliary \cite{liu2018learning} & -     & -     & 28.40 & -     & 27.60 & -     & -     & -     \\
        MADDG \cite{shao2019multi}           & 17.69 & 88.06 & 24.50 & 84.51 & 22.19 & 84.99 & 27.89 & 80.02 \\
        \hline
        \textbf{SSDG-M} & \textbf{16.67} & \textbf{90.47} & \textbf{23.11} & \textbf{85.45} & \textbf{18.21} & \textbf{94.61} & \textbf{25.17} & \textbf{81.83}\\
        \textbf{SSDG-R} & \textbf{7.38}  & \textbf{97.17} & \textbf{10.44} & \textbf{95.94} & \textbf{11.71} & \textbf{96.59} & \textbf{15.61} & \textbf{91.54} \\
        \bottomrule
    \end{tabular}
    \vspace{-0.1in}
    \label{tab:comparison_sota}
\end{table*}
\subsubsection{Limited Source Domains}
We also evaluate our method when extremely limited source domains are available (\emph{i.e.}, only two source databases).
Specifically, MSU and Idiap databases are selected as the source domains for training and the remaining two, \emph{i.e.}, CASIA and OULU, respectively, are used as the target domains for testing.
As shown in Table \ref{tab:limited_domain}, our proposed method achieves the best performance, which has a significant improvement over other methods.
Although only two source domains are available, the SSDG method can still force the features of fake faces to be dispersed in the feature space, which promotes to learn a more generalized class boundary for unseen domains.

\subsubsection{Comparisons of Different Architectures}
As shown in Table \ref{tab:backbone_comparison}, we also compare two different architectures of the feature generator, \emph{i.e.}, MADDG-based and ResNet18-based networks, respectively, to evaluate the effects of different backbones.
Specifically, the Avg HTER and AUC represent the average results of four testing tasks.
And the inference speed of each architecture is tested on the OULU database on a single NVIDIA TITAN 1080 GPU with 256$\times$256 image resolution.
It can be seen that the ResNet18-based network is more suitable than the MADDG-based one for face anti-spoofing not only in terms of accuracy but also in terms of speed.
And we believe that much better performance can be achieved by the SSDG method with more effective networks.

\subsection{Comparison with State-of-the-art Methods}
As shown in Table \ref{tab:comparison_sota} and Figure \ref{fig:roc}, our method outperforms all the state-of-the-art methods under four testing tasks, which demonstrates the effectiveness of the SSDG method.
This is because all other face anti-spoofing methods \cite{boulkenafet2016faceLBP, de2014face, liu2018learning, maatta2011face, wen2015face, yang2014learn} except for the MADDG \cite{shao2019multi} method pay no attention to the intrinsic distribution relationship among different domains.
Therefore, only database-biased features can be extracted, which causes significant performance degradation in case of cross-database testing scenarios.
Although the MADDG method exploits the DG approach to extract common discriminative cues, the results show that seeking a generalized feature space for both the real and the fake faces is difficult to optimize, usually leading to a sub-optimal solution.
Due to the diversity of attack types and database collection ways, the extracted features of fake faces are more widely distributed in the feature space than those of real ones, making it nontrivial to aggregate all of them from different domains together.
Therefore, our SSDG method applies asymmetric optimization goals for the real and the fake faces to learn a more generalized feature space.
Moreover, it shows that when we resort to using the ResNet18-based network, a significant improvement can be made, indicating that better performance can be achieved when the SSDG approach is combined with a more effective network. 

\subsection{Conclusion}
To improve the generalization ability of face anti-spoofing, we propose a novel end-to-end single-side domain generalization framework.
Our SSDG learns a generalized feature space, where the feature distribution of real faces is compact while that of fake ones is dispersed across domains.
This is quite different from existing methods treating both real and fake faces symmetrically.
To achieve this ``single-side'' goal, the single-side adversarial learning and the asymmetric triplet loss are designed to train the model aggregating the real faces and separating the fake ones from different domains. 
Extensive experiments show that our SSDG is effective and achieves state-of-the-art results on four public databases. 
In summary, our work implies that the distribution of real faces and that of fake ones are indeed different, and thus suggests that treating them asymmetrically can lead to better generalization ability to unseen domains. 
Other possible asymmetric design can be further explored in the future, for instance, dividing the fake faces according to the attack types rather than the databases. 

\section*{Acknowledgements}
This work is partially supported by National Key R\&D Program of China (No. 2017YFA0700800) and Natural Science Foundation of China (Nos. 61806188, 61772496).

{\small
\bibliographystyle{ieee_fullname}
\bibliography{references}
}

\end{document}